\documentclass{article}

\PassOptionsToPackage{numbers, compress}{natbib}
\usepackage[preprint]{neurips_2026}
\usepackage{microtype}
\usepackage{hyperref}
\usepackage{url}
\usepackage{booktabs}
\usepackage{siunitx}
\usepackage{xspace}
\usepackage{enumitem}
\usepackage{graphicx}
\usepackage{xcolor}
\usepackage{amsmath}
\usepackage{makecell}

\usepackage[utf8]{inputenc} 
\usepackage[T1]{fontenc}    
\usepackage{hyperref}       
\usepackage{url}            
\usepackage{booktabs}       
\usepackage{amsfonts}       
\usepackage{nicefrac}       
\usepackage{microtype}      
\usepackage{xcolor}         

\usepackage{lineno}

\definecolor{darkblue}{rgb}{0, 0, 0.5}
\hypersetup{colorlinks=true, citecolor=darkblue, linkcolor=darkblue, urlcolor=darkblue}
\newcommand{\ours}{Behavior Cues\xspace}
\newcommand{\bcreasoning}{Behavior Cue Reasoning\xspace}

\title{\bcreasoning: Monitorable Reasoning Improves Efficiency and Safety through Oversight}

\author{%
  Christopher Z. Cui\textsuperscript{1}\:\:\:\:Taylor W. Killian\textsuperscript{2}\:\:\:\:Prithviraj Ammanabrolu\textsuperscript{1} \\
  \textsuperscript{1}University of California, San Diego\:\:\:\:
  \textsuperscript{2}Brigham Young University\\
  \texttt{czcui@ucsd.edu}
}

\begin{document}

\maketitle

\begin{abstract}
Reasoning in Large Language Models (LLMs) poses a challenge for oversight as many misaligned behaviors do not surface until reasoning concludes. 
To address this, we introduce \bcreasoning for making LLM reasoning more controllable and monitorable.
\ours are special token sequences that a model is trained to emit immediately before specific implicit and explicit behaviors, acting as dual purpose signal and control levers.
Our experiments reveal that a \bcreasoning model has equal to greater performance as the base model, allows for steerable reasoning through external enforcement of \ours, and improves the monitorability of reasoning for external oversight monitors.
When fine-tuning a weaker external monitor with Reinforcement Learning for reasoning oversight, a compressed view of only information surfaced by \ours is sufficient signal for the monitor to prune up to 50\% of otherwise wasted reasoning tokens in complex math problem solving. 
When leveraged by an almost optimal rule-based monitor in an environment where excessive constraint violations results in failure, \ours allows for the recovery of safe actions from 80\% of reasoning traces that would otherwise end with the proposal of an unsafe action, more than doubling the success rate from 46\% to 96\%.
Through evaluation across two model families and three domains, we show that \bcreasoning improves reasoning monitorability and controllability with no cost to performance.
More broadly, our work progresses scalable oversight by demonstrating how the monitored model itself can be trained to reason more tractably to oversight. \href{https://github.com/christopherzc/behavior-cues}{Code}
\end{abstract}

\section{Introduction}\label{intro}
\begin{figure}[!tbh]
    \centering
    \includegraphics[width=\columnwidth]{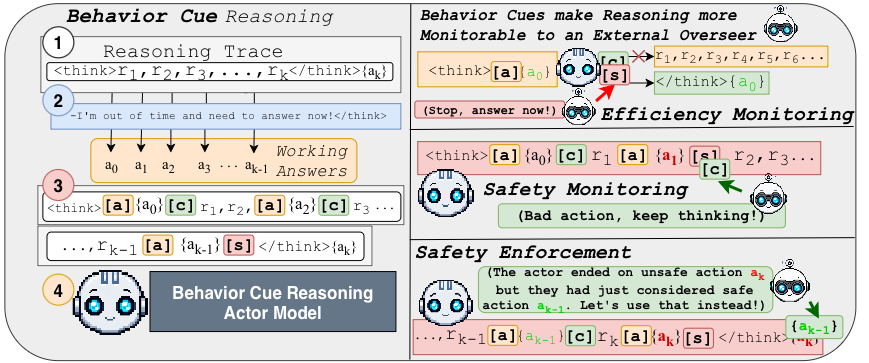}
    \caption{We take reasoning traces from a base actor model (1) and elicit the immediate working answer at each step by truncating the reasoning, appending a stop phrase, and allowing the model to produce an answer (2). Working answers are filtered to emulate the behavior of the working answer \textit{only} surfacing when changing and embedded back into the reasoning trace with \ours (3). Post SFT (4), the \bcreasoning Model adopts this answer reporting behavior, creating ideal decision points for an external oversight monitor. In \textbf{Efficiency Monitoring}, the monitor optimizes stopping once the correct final answer surfaces. In \textbf{Safety Monitoring}, the monitor attempts to prevent the commitment of unsafe actions by continuing reasoning.
    In \textbf{Safety Enforcement}, the monitor records the latest safe action to commit instead, allowing for secured unsafe reasoning.
    }
    \label{figure_1}
\end{figure}

For difficult tasks, Reasoning LLMs are trained to search within their token space by generating reasoning traces to refine their final answer.
These traces allow the model to trade inference-time compute for better performance~\citep{openai2024learning, yang2025qwen3, anthropic2025extendedthinking}, but are typically treated as artifacts excluded from the message history with the exception of tool-use~\citep{deepseekai2026deepseekv4}.
Most critically, these behaviors can remain hidden in the reasoning trace, surfacing only when the model commits to an answer.
Without deeper visibility into these reasoning traces, an external monitor tasked with oversight has limited ability to detect and interrupt these behaviors before their negative effects manifest in the model's final response.

Prior approaches either inject steering phrases into the reasoning trace at decoding time \citep{jin-etal-2025-well, yang2025testtimepromptintervention, zhang-etal-2025-steer, wu2025effectively, muennighoff-etal-2025-s1}, or train the model to condition on special tokens inserted externally into its context \citep{goyalthink, kim2025learning,  wen2025budgetthinker}. 
Some combine both, using special tokens the model is trained to respond to as steering phrases \citep{ringel-etal-2025-learning}.
In all cases, the phrase or tokens are treated as external signals.
The model itself does not surface information about its own underlying behavior, limiting applicability to oversight.

To address this, we introduce \ours to progress scalable oversight by training the model to naturally generate special token sequences that signal specific behaviors during reasoning, including behaviors that would otherwise be internal to the model. 
We define this as \bcreasoning.
The explicit appearance of \ours in a model's reasoning allows an external monitor to track behavioral progression through simple parsing, treating the model itself as a black box. 
The same tokens that surface behavior can also be externally enforced to trigger it.
This allows an external oversight monitor to intervene in a model's reasoning, extending monitorability into controllability.

In this work, we investigate three \ours: \texttt{[answer]}, \texttt{[continue]}, and \texttt{[stop]}.
We evaluate on Qwen3-8B (hybrid reasoning) and GLM-Z1-9B (pure reasoning) to test generalization across different model families in three separate problem domains.
AIME serves as a single-turn evaluation for reasoning models in complex math problem solving.
Textworld, a situated text environment with natural language observations, evaluates \ours in an out-of-domain, multi-turn, task completion setting.
Finally, we extend Hazardworld with a text-based interface to allow for evaluation of task completion under constraint adherence in multi-turn settings.


We evaluate \ours for scalable oversight through three research questions, one focused on the actor model itself and two on the monitorability of \bcreasoning to an external oversight module.
For the actor model, we evaluate \textbf{To what degree does training a model to perform \bcreasoning impact baseline task performance and controllability?}
For the monitorability of \bcreasoning, we evaluate oversight module performance along two axes: \textbf{To what degree do \ours enable an external monitor to reduce wasted reasoning tokens without sacrificing correctness} and \textbf{To what degree do \ours enable an external monitor to prevent unsafe actions?}

Our results demonstrate overall positive effects on baseline performance, adherence to external enforcement, and how \ours can enable oversight where standard reasoning would be too noisy or opaque.
We observe performance improvements in all domains and both model families.
In the majority of cases, the actor models follow injected \ours with the expected behavior.
Most significantly, \ours serve as a critical interface for external oversight.
We observe that \ours act as a necessary filter for decision points, limiting queries to the oversight module to when intervention results in the highest impact.
This provides a learnable horizon for RL fine-tuning a non-reasoning monitor.
For safety constraint enforcement, we find LLM-based monitors struggle with identifying unsafe actions. 
However, by parsing candidate actions surfaced at \texttt{[answer]} against rules constructed from training trajectories, a rule-based monitor achieves near perfect classification accuracy.
An LLM augmented with this monitor achieves 50\% more success in an environment where excessive constraint violations results in failure.
Through these experiments, we show that improved monitorability through \bcreasoning enables oversight for both safety and efficiency.

In summary, our contributions are as follows:
\begin{itemize}[itemsep=0pt, topsep=0pt, partopsep=0pt, leftmargin=*]
    \item We introduce \ours, a mechanism for making LLM reasoning more monitorable and controllable for oversight through trained token sequences (\texttt{[answer]},\texttt{[continue],[stop]}) that act as surface-level hooks for reasoning behaviors.
    \item We show a model trained to perform \bcreasoning displays overall equal to greater baseline performance and reliably adheres to expected behavior when the cue is enforced externally.
    \item For efficiency monitoring, we demonstrate that \bcreasoning enables external oversight via tractable decision horizons, allowing for the training of a non-reasoning monitor and enabling a compressed-trace formulation that correctly prunes up to 50\% of wasted reasoning tokens.
    \item For safety monitoring, we demonstrate that \bcreasoning enables a near-perfect rule-based safety monitor to recover safe actions from 80\% of otherwise unsafe reasoning traces, raising success rate from 46\% to 96\% in a safety constrained setting.
   \item We publicly release our LLM-compatible Hazardworld extension as an artifact for reasoning oversight research.
\end{itemize}

\section{Background}
\textbf{Reasoning traces.} Formally, we define a reasoning trace as the content produced by an LLM before it commits to a final answer.
For the models investigated in this work, these reasoning traces are enclosed within \texttt{<think></think>} tags. 
We define a trajectory step as the input-output pair corresponding to a single LLM call, and define a reasoning step as the portions of reasoning traces separated by two newlines.
For example, the single-turn question-answer paradigm of AIME is a single trajectory step with potentially multiple reasoning steps.
A trajectory is defined as the full sequence of trajectory steps made by the LLM when attempting a task.
We define the working answer of an LLM as the answer the LLM would provide if its reasoning were terminated at any step.\footnote{In this work, we elicit the working answer by appending \texttt{- I'm out of time and need to answer!\textbackslash n</think>}. However, we note in degenerate cases the LLM may not produce an end of sequence token after providing an answer.}
When there is a ground-truth correct answer, such as in AIME, we specifically refer to reasoning traces where the model never arrives at this correct answer as a \textbf{reasoning dead-end}.

\textbf{Scalable Oversight.}
Scalable oversight refers to the challenge of supervising AI systems whose skills or capabilities exceed that of evaluators \citep{amodei2016concrete, bowman2022measuring}. 
One research direction in this domain is the weak-to-strong paradigm, where weak overseers are studied in the context of supervising a stronger actor \citep{burns2024weaktostrong, kenton2024weakjudge}. 
Our oversight monitor experiments reside within this framing as an evaluation of the effectiveness of \ours at making an actor's reasoning more monitorable and controllable to oversight.
We discuss other works within the domain of scalable oversight in Section \ref{sec:related}.

\textbf{External Oversight.}
When evaluating the impact of \ours in making an LLM's reasoning more monitorable and controllable to oversight, we refer to the reasoning model as the \textit{actor model}.
While prior work has investigated internal model signals such as output distributions \citep{quamar2025logitentropyadaptivestoppingheuristic, sharma2025thinkjustenoughsequencelevel, yang2025dynamic} or hidden state representations \citep{azaria2023internal, zhang2025probinghidden}, we treat the actor model as a black box where only the generated tokens are observed.
In AIME and Cookingworld, the oversight module is restricted to non-reasoning mode and constrained to only outputting \texttt{[continue]} or \texttt{[stop]}.
At each decision point, the oversight module observes the actor model's reasoning and context and decides whether further reasoning is needed.
In Hazardworld, we evaluate two oversight approaches under the same no-reasoning-budget constraint: an LLM-based monitor restricted to 'safe' or 'unsafe' outputs, and a rule-based monitor that specifically parses the latest candidate action surfaced by \texttt{[answer]}.


\section{\ours and External Oversight}\label{meth}

\textbf{Behavior Cues.} We define \ours as special token sequences an LLM is trained to naturally emit during reasoning immediately before specific behaviors.
In this work, we explore the application of three \ours:
\begin{itemize}[itemsep=0pt, topsep=0pt, partopsep=0pt, leftmargin=*]
    \item \texttt{[answer]}: updates the working answer. We explicitly train the model to provide its best-guess answer without reasoning, resulting in a progression of answers throughout the reasoning phase.  
    \item \texttt{[continue]}: after a new working answer, \texttt{[continue]} signals further reasoning token generation.
    \item \texttt{[stop]}: after a new working answer, \texttt{[stop]} signals a termination of reasoning.
\end{itemize}
Thus, when a model trained to produce these \ours is generating the answer to a question, \texttt{[answer]} should always appear at the start of reasoning immediately after the \texttt{<think>} tag and be followed by what is functionally the model's non-reasoning best guess answer. 
If the model then generates \texttt{[continue]}, it should proceed with its reasoning and produce another \texttt{[answer]} when its working answer changes.
If the model generates \texttt{[stop]}, it should immediately stop its reasoning.
These \ours thus are meant to serve a dual purpose of both signaling the occurrence of specific behaviors as well as acting as enforceable levers.
Figure \ref{figure_1} shows how the actor model is trained to naturally produce \ours as well as how they can be used by an external module as a sample efficient interface for oversight.
Attempts to elicit \bcreasoning through prompting or RL were both unsuccessful. See Appendix \ref{app:ans_fix} for more details.


Let $r_0, r_1, r_2, ..., r_{k-1}$ represent the steps in a model's reasoning and $a_0, a_1, a_2, ..., a_{k-1}$ represent the model's working answers for each corresponding reasoning step.
We train a model to perform \bcreasoning through an elicitation, embedding, and SFT pipeline.
First we elicit the working answer at each reasoning step for every task in the training set by truncating reasoning, appending a termination phrase, and allowing the model to produce a final answer\footnote{This approach is similar to what is done by Mao et al.~\citep{mao2025earlystoppingchainofthoughtslarge}}. 
Then, these working answers are embedded into the reasoning between \texttt{[answer]} and \texttt{[continue]} (or \texttt{[stop]}) blocks to create the SFT dataset.
At step $r_n$, the answer $a_n$ is only embedded if it is not equal to the last embedded answer.
This ensures each occurrence of \texttt{[answer]} is meaningful, marking a change in the model's most probable answer.
To isolate the effects of any performance drift to training a model to perform \bcreasoning specifically, we deliberately do not filter reasoning traces for correctness of the final answer.

\textbf{Oversight Module.} The oversight module is a separate monitor that tracks the actor model's context and reasoning progression and can intervene at specific decision points.
Both actor and monitor are treated as mutually opaque: neither has access to the other's internals.
In deployment, the monitor's primary means for steering the actor model is through token forcing at decoding time.
The monitor is trained offline on pre-generated traces from a \bcreasoning model to allow for ground-truth verification of impact.
We use the \bcreasoning actor model to generate reasoning traces instead of assuming the perfect adherence that would arise from using the base model's traces and completions.
These reasoning traces are then split 80-20 into training and validation.
We use the Qwen3-8B model set to ``no-think mode'' as a base for fine-tuning.

The occurrence of \texttt{[answer]} can be used as an information dense signal for when to query the oversight module, filtering these queries to only when the actor model's answer actually changes.
Due to the length of reasoning for complex problems, it can be inefficient to query the monitor in a fine-grained manner such as after every reasoning step.
While querying after every reasoning step allows for finer-grained intervention, it introduces substantial noise during training from stretches where the working answer is unchanged.

We evaluate the effectiveness of \texttt{[answer]} as an information dense intervention point in two training paradigms, RL for efficiency monitoring and SFT for safety monitoring.
As the outcome is based on the output of the monitor across all decision points at the time of termination, we use RL with a terminal reward to train the oversight module for the efficiency objective.
For safety monitoring, the safety of any proposed action within the reasoning is independent of previous action classifications. 
Thus we use SFT to train the oversight module for the safety objective.

\section{Experimental Set Up}\label{sec:exp}
We evaluate \ours across two model families (Qwen3-8B and GLM-Z1-9B), and three domains, (AIME, Textworld, and Hazardworld).
We use Qwen3 as a hybrid reasoning model, trained to both reason and immediately answer, and GLM as a pure reasoning model to evaluate the transferability of our method across architectures and training regimes. 
We use AIME, Textworld, and Hazardworld to evaluate \ours in making an actor model's reasoning trace more monitorable and controllable to oversight across a diverse set of domains.
Our experiments investigate each stage at which \ours take effect from the base model itself to their impact through an external oversight monitor. For additional details, see Appendix Section \ref{app:cooking_filter} for Cookingworld, Section \ref{app:hazard_changes} for Hazardworld, Section \ref{app:RL} for RL fine-tuning, and Section \ref{app:SFT} for Supervised fine-tuning.

\textbf{AIME: }
The AIME (American Invitational Math Examination) problem set is a series of complex math problems.
We use the 2025 AIME question set as an evaluation split to minimize potential data contamination with the pre-2025 AIME question sets acting as our training set for SFT.
For AIME reasoning traces, the monitor has two objectives.
For successful reasoning traces, the monitor is tasked with stopping reasoning as soon as the correct answer first surfaces.
For dead-end reasoning traces, the monitor is tasked with stopping reasoning as soon as possible.



\textbf{Cookingworld: }
Textworld \cite{cote2018textworld} is a situated text simulator originally designed for training classical RL agents in natural language environments.
For our experiments, we use the Cookingworld scenarios used for evaluation in \cite{cui2025talestextadventurelearning} with alternate seeds and layouts as our train split. 
Cookingworld tasks involve the LLM acting as an agent to navigate a household setting, gather ingredients, and prepare a meal.
We avoid labeling reasoning traces in Cookingworld as reasoning dead-ends as due to the multi-step nature of the task, there is almost no ground-truth for if an action is 'incorrect' in the immediate context as it may lead to success or failure later on.
Please see Appendix \ref{app:cooking_filter} for details.

\textbf{Hazardworld: }
Hazardworld \cite{yang2021safe} is a Minigrid environment originally designed for training RL agents.
In Hazardworld, the agent must navigate to a series of objects while avoiding a set of randomly generated hazard tiles.
The specific type of hazard tile the agent must avoid is provided in a natural language description such as \texttt{Avoid Water} or \texttt{Stay at least 1 block away from Lava}.
Unsafe actions in Hazardworld violate the provided constraints.
We report the adjusted average reward, where the LLM receives 0 reward if the total constraint violation budget is exceeded.
For our experiments, we extend Hazardworld to be compatible with LLM agents.
See Appendix \ref{app:hazard_changes} for a comprehensive list of modifications to the original environment.
In Hazardworld, the monitor is tasked solely with the classification of actions as safe or unsafe given the current actor model context.




\section{Evaluation}\label{results}
We center our evaluation around three research questions investigating the actor model and oversight module.
These questions evaluate the actor model across downstream performance and controllability, and the \bcreasoning traces themselves in how they lend themselves to better enabling an external oversight module to perform efficiency monitoring and safety monitoring. 

\textbf{RQ1: To what degree does training a model to perform \bcreasoning impact baseline task performance and controllability?} 
Methods for improving the controllability or monitorability of models typically come at the cost of some negative impact on performance.
In these initial experiments, we examine the degree of performance drift from fine-tuning an actor model to perform \bcreasoning as well as how controllable the actor model is to externally enforced \ours.
To evaluate task performance, we fix decoding parameters per domain and report performance on a held-out test set.
To evaluate \ours controllability, we randomly sample 200 decision points for each \bcreasoning model where the model would output \texttt{[continue]} or \texttt{[stop]} (100 each), replace it with the opposite reasoning cue, and allow the model to continue generating.
We measure the rate of the actor model adhering to the replaced cue.
Adherence to \texttt{[continue]} is calculated as the number of times the model generates at least one or more \texttt{[answer]} blocks before terminating reasoning.
Adherence to \texttt{[stop]} is calculated as the number of times the model immediately generates the \texttt{</think>} tag and reports the last reported answer.

\begin{table}[!tbh]
\centering
\caption{Top: task performance. Bottom: adherence to forced reasoning cues (continue / stop), for the Behavior Cues variants. The Hazardworld \textbf{adjusted score} is calculated on the basis of a 0 overall reward for trajectories where the constraint violation budget was exceeded. See Table \ref{tab:monitor_results} for a more detailed breakdown. We see overall performance gains across the board for all models.}
\renewcommand{\arraystretch}{1.2}
\setlength{\tabcolsep}{8pt}
\begin{tabular}{l S[table-format=2.1] S[table-format=2.1] S[table-format=1.1]}
\toprule
& {\textbf{AIME2025}} & {\textbf{Cookingworld}} & {\textbf{Hazardworld}} \\
\textbf{Model} & {(\% Correct)} & {(\% Wins)} & {(Adj.\ Score)} \\
\midrule
Qwen3-8B                   & 62.5          & 53.3          & \bfseries 2.3 \\
Qwen3-8B w/ Behavior Cues  & \bfseries 69.2 & \bfseries 56.7 & 2.4          \\
GLM-Z1-9B                  & 54.4          & 36.6          & 1.3          \\
GLM-Z1-9B w/ Behavior Cues & 56.7          & 46.7          & 1.3          \\
\midrule
& \multicolumn{3}{c}{\textit{Adherence to forced reasoning cues (\% continue / \% stop)}} \\
\midrule
Qwen3-8B  & {30.9 / 100.0} & {84.0 / 55.0} & {97.0 / 100.0} \\
GLM-Z1-9B & {34.4 / 100.0} & {75.5 / 63.5} & {87.0 / \phantom{0}98.0} \\
\bottomrule
\end{tabular}
\label{tab:math_tw_hw_results_compact}
\end{table}

\textit{\textbf{Training a model to perform \bcreasoning has minimal to positive effects on baseline task performance while adherence to externally enforced \ours differs by domain.}}
Table \ref{tab:math_tw_hw_results_compact} shows the \bcreasoning model matches or surpasses base model performance across all domains.
The bottom half of Table \ref{tab:math_tw_hw_results_compact} shows the adherence to externally enforced \texttt{[continue]} and \texttt{[stop]}.
While we see overall adherence, there is a clear stratification by domain.
In Hazardworld, reasoning traces can be long (up to 8k tokens) but the model sees many examples of \texttt{[proposed action]} followed by \texttt{[stop]} due to the smaller action space.
In contrast, reasoning traces from Cookingworld are overall far shorter (up to 2k tokens) and typically involve a similar set of actions. 
As a result, the model learns the \texttt{[stop]} behavior to be associated with a specific subset of actions and reasoning length.
In particular, we note adherence to \texttt{[stop]} has an inverse relationship to the reasoning length with practically zero adherence when injected at the very first \texttt{[answer]} block, rapidly increasing as reasoning continues and switches between more potential actions.
AIME shows the other extreme with an unbounded answer space and highly variable length reasoning traces.
Only \texttt{[stop]} is learned with the specific behavior we intended.
In AIME specifically, we note that once a high-confidence answer is surfaced, the model never deviates, and reasoning quickly terminates again on the same answer if forced to continue. 
Notably, we observe \textbf{no answer changes} even when Qwen adheres to \texttt{[continue]}, and only a 16\% answer-change rate for GLM.
Across all \bcreasoning traces for both models, we observe a \textbf{less than 3\% rate} of the model proposing the correct answer only to later change to another answer.

\textbf{RQ2: To what degree do \ours enable an external monitor to reduce wasted reasoning tokens without sacrificing correctness?}
To evaluate the monitorability of \bcreasoning versus standard reasoning in the efficiency setting, we train the oversight module described in Section \ref{meth} in an offline setting where reasoning traces are pre-generated.
The offline setting enables a ground-truth comparison of the final committed answer versus the answer stopped at by the monitor.
This enables calculation for overall tokens saved while preserving correctness of the final solution or tokens saved through early termination of reasoning dead-ends.
We compare two monitor-training formulations enabled by \ours: using \texttt{[answer]} occurrences as decision points versus using individual reasoning steps.
We additionally evaluate whether the information surfaced by \ours alone is sufficient signal for the monitor by training on a compressed view of the reasoning traces containing only the answer progression and the number of reasoning lines between each surfaced answer.
AIME's long reasoning traces make this compressed-trace setting informative for testing whether \ours alone exposes sufficient signal for early reasoning termination.

\begin{figure}[!tbh]
    \centering
    \includegraphics[width=\columnwidth]{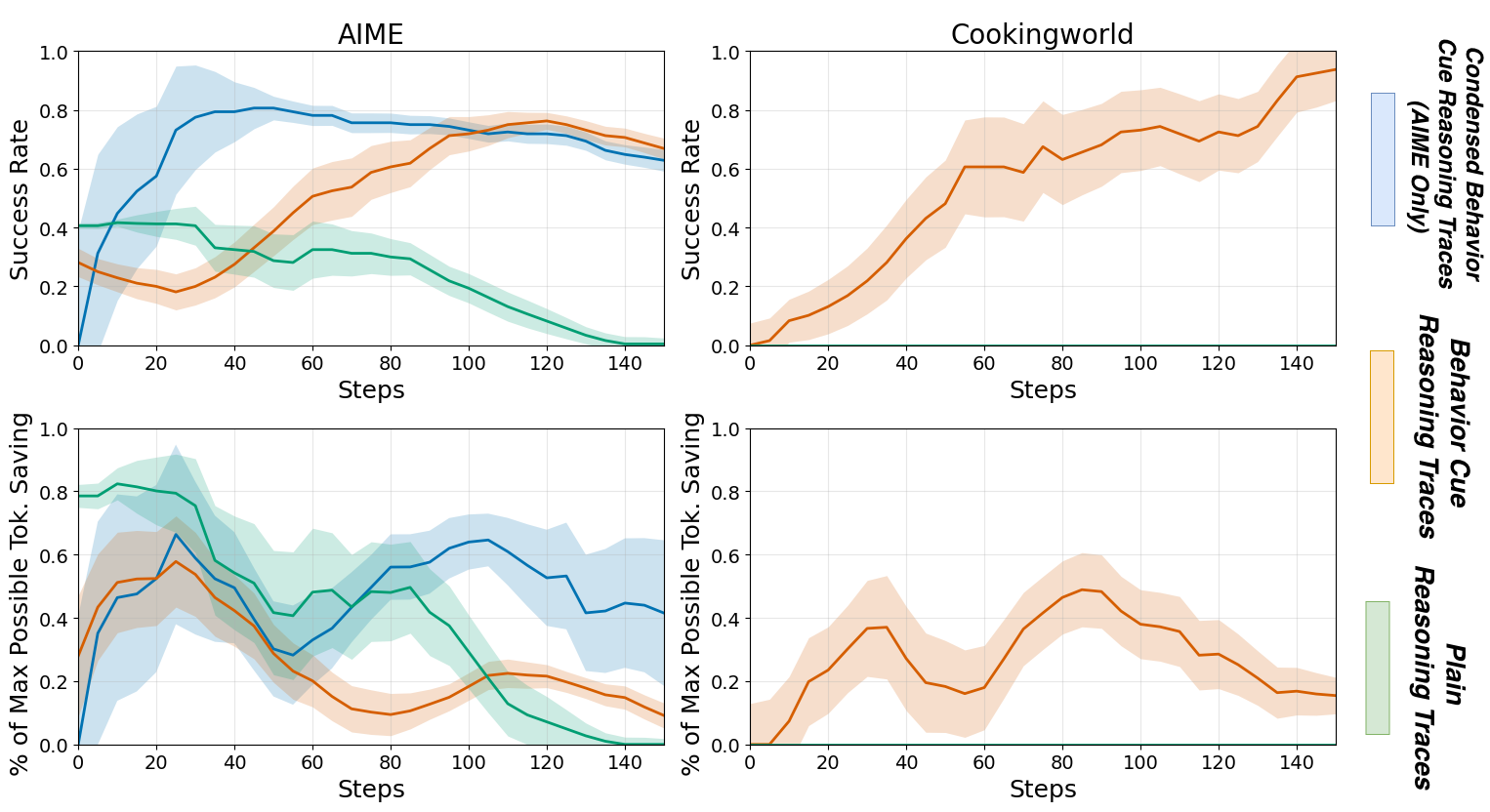}
    \caption{Validation success rate and percent of maximum possible token savings over the course of RL fine-tuning for the oversight module. In AIME, due to the single-turn nature and reasoning trace length, we include a secondary objective of terminating dead-end reasoning traces as early as possible. Additionally, due to the length of these reasoning traces, we include a monitor training setting where the actor's reasoning is compacted: a formulation enabled solely due to \ours. In Cookingworld, we find the plain reasoning trace results in the monitor finding no success or reward and quickly collapsing to a degenerate policy.}
    \label{fig:monitor_rl}
\end{figure}

\textit{\textbf{\ours serve a critical role in allowing a non-reasoning oversight module to learn to supervise a reasoning actor model for efficiency.}}
While using individual reasoning steps as decision points allows for more fine-grained intervention, we find it quickly diverges and leads to degenerate policies during training.
Figure \ref{fig:monitor_rl} shows success rate and percentage of ground-truth token savings achieved during RL training for monitors in both AIME and Cookingworld.
For AIME specifically, we include the results from an ablation where the reasoning trace is compressed to \textit{only} the answer progression and reasoning line count between answer switches. 
We observe for \bcreasoning Traces, the monitor begins to calibrate decisions on when to terminate reasoning based on the actor model's reasoning itself. 
While effective for pruning excess tokens from correct trajectories, this policy is far less effective at pruning dead-end reasoning, shown by the lower achieved percentage of possible token savings in AIME.
The monitor in these scenarios learns to determine stopping criteria by effectively internalizing the actor model's reasoning and stopping just short of when the actor model's natural reasoning would stop.
The compressed reasoning prevent this learning shortcut by stripping away the reasoning content itself, forcing the monitor to learn to stop based on answer progression patterns during reasoning instead.
This suggests the answer progression in reasoning traces leading to correct answers is generally distinct from ineffective reasoning that ultimately fails.

\textbf{RQ3: To what degree do \ours enable an external monitor to prevent unsafe actions?}
To evaluate the monitorability of \bcreasoning versus standard reasoning in the safety setting, we perform two experiments. 
Similar to the efficiency setting, we evaluate the degree to which \bcreasoning traces are more tractable versus plain reasoning traces for a non-reasoning monitor learning to classify reasoning as leading to safe or unsafe actions.
Next, we evaluate a safety-enforcement setting that \bcreasoning enables.
A rule-based action classification monitor is constructed purely from action outcomes in training trajectories.
This rule-based monitor is paired with a \bcreasoning actor model to record safe candidate actions surfaced via \texttt{[answer]} during reasoning, substituting them for any unsafe action the actor's reasoning ultimately concludes in.
This isolates the contribution of \ours to the recovery mechanism: while the classifier does not require \ours, the mid-reasoning extraction and substitution of safe candidate actions is not possible with standard reasoning.
We evaluate this setting online with the deployed monitor to measure downstream effects on success that offline evaluation could not capture.

\begin{table}[!tbh]
\centering
\small
\caption{F1 scores from Supervised Fine-tuning on classifying actions as `Safe' or `Unsafe' in Hazardworld. We report F1 over accuracy due to class imbalance in the action distribution. Both settings struggle, with the monitor having a slightly easier time on \bcreasoning Traces.}
\begin{tabular*}{\linewidth}{@{\extracolsep{\fill}}lrrrrrrrrrrr}
\toprule
\textit{Reasoning Format} & \multicolumn{11}{c}{Input Tokens Trained On (Million)} \\
\cmidrule(lr){2-12}
                  & Base & 1M & 2M & 3M & 4M & 5M & 6M & 7M & 8M & 9M & 10M \\
\midrule
\ours             & 58\% & 12\% & 60\% & 33\% & 63\% & 55\% & 36\% & 45\% & 66\% & 60\% & 60\% \\
Plain Reasoning   & 39\% & 40\% & 39\% & 35\% & 43\% & 41\% & 31\% & 22\% & 30\% & 49\% & 48\% \\
\bottomrule
\end{tabular*}
\label{tab:monitor_sft}
\end{table}

\textit{\textbf{LLM-based safety classification is difficult even with \ours.}}
Table \ref{tab:monitor_sft} reports the F1 of an SFT-trained non-reasoning monitor classifying stretches of reasoning as leading to safe or unsafe actions. 
\bcreasoning traces are compared against plain reasoning.
While \bcreasoning sees a generally higher F1 across most checkpoints, neither setting reaches the reliability needed for a deployable safety monitor.
This finding suggests the difficulty in this setting stems from the inability of the base LLM itself to perform the safety classification without reasoning.
This motivates an alternative approach: rather than improve LLM-based classification, we leverage \ours to enable an intervention that does not require an LLM-based monitor at all.

\textit{\textbf{\bcreasoning enables recovery of safe actions from reasoning that ultimately ends unsafely, drastically boosting success rate.}}
The near-perfect accuracy of the rule-based monitor (99.54\% F1) allows us to explicitly isolate the contribution of \ours from monitor accuracy.
Table \ref{tab:monitor_results} demonstrates the effect of pairing a near-optimal rule-based monitor with a \bcreasoning model in the recovery setting, more than doubling overall success rate from 46\% to 96\%.  
\texttt{[answer]} enables a reasoning-preserving mode that a simple action reject monitor is unable to do.
Of all reasoning traces that ultimately terminate on an unsafe action, over 80\% have a considered safe action recovered by the monitor through \texttt{[answer]}.
Rather than throw out all reasoning and force the actor model to generate an entirely new reasoning trace, the safety enforcement monitor is able to extract the model's most confident safe action exposed by \texttt{[answer]}, despite the reasoning ultimately ending in an unsafe action.

\begin{table}[h]
\centering
\caption{Adjusted reward zeros out the score for any episode where violations exceed the budget. The Enforcement Monitor prevents the commitment of any unsafe action, falling back to a no-op action or the last proposed safe action during reasoning.}
\label{tab:monitor_results}
\begin{tabular}{lccc}
\toprule
Configuration & Success \% & Adj.\ Reward & Avg.\ Reward \\
\midrule
Qwen3-8B                                  & 46\% & 2.35 & 4.86 \\
Qwen3-8B w/ Behavior Cues                   & 46\% & 2.45 & 4.64 \\
Safety Enforcement Monitor      & \textbf{96\%} & \textbf{3.99} & 3.99 \\
\bottomrule
\end{tabular}
\end{table}

\section{Related Works}\label{sec:related}
\textbf{Reasoning Monitorability and Oversight:}
Recent work examines the potential of monitoring LLM reasoning to detect and address misaligned behaviors before they surface in the final response \citep{korbak2025chainthoughtmonitorabilitynew, bowman2022measuring, guan2025monitoringmonitorability}.
Previous work has shown that these reasoning traces can still be unfaithful, where the final answer of the model doesn't align with the reasoning \citep{turpin2023languagemodelsdontsay, lanham2023measuringfaithfulnesschainofthoughtreasoning, howe2026endsjustifythoughtsrlinduced, greenblatt2024alignmentfakinglargelanguage}.
This unfaithfulness becomes an issue when an overseer needs to infer the actor model's working answer from the reasoning, as unfaithfulness may reduce the monitor's precision in identifying the actor's current working answer.
Baker et al.\citep{baker2025monitoringreasoningmodelsmisbehavior} demonstrate that while reasoning monitors can detect misaligned behaviors in frontier reasoning models, applying optimization pressure directly to the reasoning trace can produce obfuscated misbehavior, where the model learns to hide misaligned behavior rather than avoid them to improve reward: this motivates a "monitorability tax", where reasoning is deliberately left outside the training loop to explicitly isolate it from optimization pressures.
\ours aligns with this principle and side steps the issue of faithfulness by training an actor model to directly expose its working answer at meaningful junctions.
All optimization pressure for oversight is instead placed on the monitor and interpreting the actor model's reasoning becomes a non-issue.

\textbf{Reasoning Control through Token Injection:} 
Prior work examines the application of interventions for steering a model's reasoning.
These approaches typically involve the injection of natural language phrases to steer the model towards certain thought paths without modification of the underlying model~\cite{jin-etal-2025-well, yang2025testtimepromptintervention, zhang-etal-2025-steer, wu2025effectively, tang2025concisehint}.
\citep{muennighoff-etal-2025-s1} train a non-thinking model to reason effectively, showing performance can be further improved by early termination or the use of the phrase "Wait" to extend reasoning.
\citep{zhao2025activation} leverage activations and the "wait" token to elicit long chain of thought behavior with minimal training.
In contrast to these methods, our approach specifically targets making a model's reasoning trace more monitorable \textbf{and} steerable.

\textbf{Inducing Behavior Through Control Tokens:}
Previous works have explored the use of controlling model behavior via special tokens, both inside and outside of a reasoning trace.
\citep{zhangbacktracking} leverage the [RESET] token to allow a model to completely undo potentially unsafe generations with \citep{sel2025backtracking} refining this to a specific subset of the generation.
More recent work extend the use of special tokens to reset behavior to reasoning~\citep{yang2025step}.
\citep{ringel-etal-2025-learning} demonstrate how a specialized token for enforcing continued reasoning can outperform a natural language signal such as "Wait".
\citep{kim2025learning} and \citep{goyalthink} use the [PAUSE] token albeit in different ways, the former allows the model to express low confidence and the latter to signal the model to perform more internal hidden layer calculations, effectively increasing the compute used for generating the next token.

\textbf{Early Stopping in Reasoning:}
In this section, we discuss work that explicitly terminate or truncate reasoning early for both better token efficiency and performance.
These approaches typically rely on entropy, or the confidence in a proposed answer as a signal to immediately terminate the model's reasoning at a certain reasoning step~\citep{yang2025dynamic, sharma2025thinkjustenoughsequencelevel, quamar2025logitentropyadaptivestoppingheuristic}.
\citep{sun2025stopenoughadaptiveearlystopping} uses the similarity of newly generated reasoning steps with previous steps as a signal for early termination.
\citep{mao2025earlystoppingchainofthoughtslarge} generates answers for each reasoning step, ending reasoning when the answer is consistent through a large number of steps. 

\textbf{Reasoning Agents in Interactive Tasks}
Agents in interactive environments have been studied long before the advent of LLMs with language-focused testbeds such as Textworld \cite{cote2018textworld} or Jericho \cite{hausknecht2020interactive} used for training RL agents in sequential decision-making \cite{shridharalfworld, wang2022scienceworld}. 
These agents often used a set of action templates to bound natural language space to a subset learnable by a smaller network.
Recent work has examined the performance of LLMs as the decision making centers of agents in these environments~\cite{cui2025talestextadventurelearning, paglieri2024balrog}. 
LLMs as agents in interactive tasks have also been explored in a wide range of domains such as coding \cite{yuan2025debuggymtextbasedenvironmentinteractive, yang2024swe}, scientific discovery \cite{nathani2025mlgym, jansen2024discoveryworld}, and deep research \cite{huang2025deep}. 

\section{Considerations}\label{sec:consider}
\textbf{Limitations.} Our work progresses scalable oversight through the explicit training of a model's reasoning to be more monitorable and controllable to external intervention.
While we attempt to minimize downstream performance impact, there is a clear distribution shift.
Regardless of whether this ultimately improves performance, future work may investigate methods for teaching a model to perform \bcreasoning without needing to modify the original model's weights at all.
We primarily investigate domains where the final answer is a short phrase that can easily be integrated into the reasoning trace.
Another direction may be pushing the model to specifically record this answer in an external document to facilitate extension to environments where the downstream answer may be longer than the reasoning itself.
Lastly, we primarily investigate the trio of \texttt{[answer]}, \texttt{[continue]}, and \texttt{[stop]} due to the implicit relation between all of these behaviors.
Future work may investigate other reasoning behaviors such as backtracking for verification.

\textbf{Societal Impacts.} Behavior Cues are intended to improve scalable oversight by making intermediate reasoning states easier to monitor and intervene on, potentially reducing wasted inference compute and preventing unsafe commitments before they appear in final outputs.
However, because many users interact with LLMs through APIs or mediated interfaces, cue-based control could be hidden from end users and used by a platform or intermediary to silently steer model reasoning without meaningful transparency or consent.
Behavior Cues should also be understood as strong learned priors rather than deterministic guarantees due the stochasticity of LLM generation, reflected by the varying degrees of cue compliance observed across domains. See Appendix \ref{app:extended} for further discussion.

\section{Conclusion}
In this work we introduce \ours, a method for making a model's reasoning more monitorable and controllable.
During \bcreasoning, the model periodically emits special token sequences that are anchored to specific behaviors: \texttt{[answer]} when the model's internal working answer changes, \texttt{[continue]} when the model believes it needs to continue reasoning, and \texttt{[stop]} when the model commits its answer.
Across two model families and three domains, we demonstrate that training an actor model to produce these \ours matches or improves baseline performance across nearly all model-domain combinations.
During \bcreasoning, the actor model will reliably adhere to externally enforced \texttt{[continue]} or \texttt{[stop]} cues while exposing its answer progression via the \texttt{[answer]}.

This answer progression enables direct oversight over the actor model's answer before it is committed, allowing for improvements in both reasoning efficiency and safety in constraint adherence.  
In reasoning efficiency, this answer progression allows a weaker, non-reasoning monitor to intervene when the actor model has already arrived at the correct answer while also preemptively pruning reasoning dead-ends, saving up to 50\% of otherwise wasted tokens.
In safety, the exposure of the actor model's working answer allows for the extraction of safe actions from reasoning traces that ultimately conclude in unsafe actions, recovering 80\% of LLM steps that might have otherwise required resampling the model.

\bibliographystyle{plain} 
\bibliography{neurips2026/neurips}

\begin{thebibliography}{10}

\bibitem{amodei2016concrete}
Dario Amodei, Chris Olah, Jacob Steinhardt, Paul Christiano, John Schulman, and Dan Man{\'e}.
\newblock Concrete problems in ai safety.
\newblock {\em arXiv preprint arXiv:1606.06565}, 2016.

\bibitem{anthropic2025extendedthinking}
{Anthropic}.
\newblock Building with extended thinking.
\newblock \url{https://docs.claude.com/en/docs/build-with-claude/extended-thinking}, 2025.

\bibitem{azaria2023internal}
Amos Azaria and Tom Mitchell.
\newblock The internal state of an {LLM} knows when it's lying.
\newblock In {\em Findings of the Association for Computational Linguistics: EMNLP 2023}, 2023.

\bibitem{baker2025monitoringreasoningmodelsmisbehavior}
Bowen Baker, Joost Huizinga, Leo Gao, Zehao Dou, Melody~Y. Guan, Aleksander Madry, Wojciech Zaremba, Jakub Pachocki, and David Farhi.
\newblock Monitoring reasoning models for misbehavior and the risks of promoting obfuscation, 2025.

\bibitem{bowman2022measuring}
Samuel~R. Bowman, Jeeyoon Hyun, Ethan Perez, Edwin Chen, Craig Pettit, Scott Heiner, Kamil{\.e} Luko{\v{s}}i{\=u}t{\.e}, Amanda Askell, Andy Jones, Anna Chen, et~al.
\newblock Measuring progress on scalable oversight for large language models.
\newblock {\em arXiv preprint arXiv:2211.03540}, 2022.

\bibitem{burns2024weaktostrong}
Collin Burns, Pavel Izmailov, Jan~Hendrik Kirchner, Bowen Baker, Leo Gao, Leopold Aschenbrenner, Yining Chen, Adrien Ecoffet, Manas Joglekar, Jan Leike, Ilya Sutskever, and Jeff Wu.
\newblock Weak-to-strong generalization: Eliciting strong capabilities with weak supervision.
\newblock In {\em International Conference on Machine Learning}, 2024.

\bibitem{cote2018textworld}
Marc-Alexandre C{\^o}t{\'e}, Akos K{\'a}d{\'a}r, Xingdi Yuan, Ben Kybartas, Tavian Barnes, Emery Fine, James Moore, Matthew Hausknecht, Layla El~Asri, Mahmoud Adada, et~al.
\newblock Textworld: A learning environment for text-based games.
\newblock In {\em Workshop on Computer Games}, pages 41--75. Springer, 2018.

\bibitem{cui2025talestextadventurelearning}
Christopher~Zhang Cui, Xingdi Yuan, Ziang Xiao, Prithviraj Ammanabrolu, and Marc-Alexandre Côté.
\newblock Tales: Text adventure learning environment suite, 2025.

\bibitem{deepseekai2026deepseekv4}
DeepSeek-AI.
\newblock Deepseek-v4: Towards highly efficient million-token context intelligence, 2026.

\bibitem{feng2025group}
Lang Feng, Zhenghai Xue, Tingcong Liu, and Bo~An.
\newblock Group-in-group policy optimization for llm agent training.
\newblock {\em arXiv preprint arXiv:2505.10978}, 2025.

\bibitem{goyalthink}
Sachin Goyal, Ziwei Ji, Ankit~Singh Rawat, Aditya~Krishna Menon, Sanjiv Kumar, and Vaishnavh Nagarajan.
\newblock Think before you speak: Training language models with pause tokens.
\newblock In {\em The Twelfth International Conference on Learning Representations}.

\bibitem{greenblatt2024alignmentfakinglargelanguage}
Ryan Greenblatt, Carson Denison, Benjamin Wright, Fabien Roger, Monte MacDiarmid, Sam Marks, Johannes Treutlein, Tim Belonax, Jack Chen, David Duvenaud, Akbir Khan, Julian Michael, Sören Mindermann, Ethan Perez, Linda Petrini, Jonathan Uesato, Jared Kaplan, Buck Shlegeris, Samuel~R. Bowman, and Evan Hubinger.
\newblock Alignment faking in large language models, 2024.

\bibitem{guan2025monitoringmonitorability}
Melody~Y. Guan, Miles Wang, Micah Carroll, Zehao Dou, Annie~Y. Wei, Marcus Williams, Benjamin Arnav, Joost Huizinga, Ian Kivlichan, Mia Glaese, Jakub Pachocki, and Bowen Baker.
\newblock Monitoring monitorability, 2025.

\bibitem{hausknecht2020interactive}
Matthew Hausknecht, Prithviraj Ammanabrolu, Marc-Alexandre C{\^o}t{\'e}, and Xingdi Yuan.
\newblock Interactive fiction games: A colossal adventure.
\newblock In {\em Proceedings of the AAAI Conference on Artificial Intelligence}, volume~34, pages 7903--7910, 2020.

\bibitem{herel2024collapseselftrainedlanguagemodels}
David Herel and Tomas Mikolov.
\newblock Collapse of self-trained language models, 2024.

\bibitem{howe2026endsjustifythoughtsrlinduced}
Nikolaus Howe and Micah Carroll.
\newblock The ends justify the thoughts: Rl-induced motivated reasoning in llm cots, 2026.

\bibitem{huang2025deep}
Yuxuan Huang, Yihang Chen, Haozheng Zhang, Kang Li, Huichi Zhou, Meng Fang, Linyi Yang, Xiaoguang Li, Lifeng Shang, Songcen Xu, et~al.
\newblock Deep research agents: A systematic examination and roadmap.
\newblock {\em arXiv preprint arXiv:2506.18096}, 2025.

\bibitem{jansen2024discoveryworld}
Peter Jansen, Marc-Alexandre C{\^o}t{\'e}, Tushar Khot, Erin Bransom, Bhavana Dalvi~Mishra, Bodhisattwa~Prasad Majumder, Oyvind Tafjord, and Peter Clark.
\newblock Discoveryworld: A virtual environment for developing and evaluating automated scientific discovery agents.
\newblock {\em Advances in Neural Information Processing Systems}, 37:10088--10116, 2024.

\bibitem{jin-etal-2025-well}
Hyunbin Jin, Je~Won Yeom, Seunghyun Bae, and Taesup Kim.
\newblock ``well, keep thinking'': Enhancing {LLM} reasoning with adaptive injection decoding.
\newblock In Wanxiang Che, Joyce Nabende, Ekaterina Shutova, and Mohammad~Taher Pilehvar, editors, {\em Findings of the Association for Computational Linguistics: ACL 2025}, pages 9989--10018, Vienna, Austria, July 2025. Association for Computational Linguistics.

\bibitem{kenton2024weakjudge}
Zachary Kenton, Noah~Y. Siegel, J{\'a}nos Kram{\'a}r, Jonah Brown-Cohen, Samuel Albanie, Jannis Bulian, Rishabh Agarwal, David Lindner, Yunhao Tang, Noah~D. Goodman, and Rohin Shah.
\newblock On scalable oversight with weak llms judging strong llms.
\newblock {\em arXiv preprint arXiv:2407.04622}, 2024.

\bibitem{kim2025learning}
Eunki Kim, Sangryul Kim, and James Thorne.
\newblock Learning to insert [pause] tokens for better reasoning.
\newblock {\em arXiv preprint arXiv:2506.03616}, 2025.

\bibitem{korbak2025chainthoughtmonitorabilitynew}
Tomek Korbak, Mikita Balesni, Elizabeth Barnes, Yoshua Bengio, Joe Benton, Joseph Bloom, Mark Chen, Alan Cooney, Allan Dafoe, Anca Dragan, Scott Emmons, Owain Evans, David Farhi, Ryan Greenblatt, Dan Hendrycks, Marius Hobbhahn, Evan Hubinger, Geoffrey Irving, Erik Jenner, Daniel Kokotajlo, Victoria Krakovna, Shane Legg, David Lindner, David Luan, Aleksander Mądry, Julian Michael, Neel Nanda, Dave Orr, Jakub Pachocki, Ethan Perez, Mary Phuong, Fabien Roger, Joshua Saxe, Buck Shlegeris, Martín Soto, Eric Steinberger, Jasmine Wang, Wojciech Zaremba, Bowen Baker, Rohin Shah, and Vlad Mikulik.
\newblock Chain of thought monitorability: A new and fragile opportunity for ai safety, 2025.

\bibitem{lanham2023measuringfaithfulnesschainofthoughtreasoning}
Tamera Lanham, Anna Chen, Ansh Radhakrishnan, Benoit Steiner, Carson Denison, Danny Hernandez, Dustin Li, Esin Durmus, Evan Hubinger, Jackson Kernion, Kamilė Lukošiūtė, Karina Nguyen, Newton Cheng, Nicholas Joseph, Nicholas Schiefer, Oliver Rausch, Robin Larson, Sam McCandlish, Sandipan Kundu, Saurav Kadavath, Shannon Yang, Thomas Henighan, Timothy Maxwell, Timothy Telleen-Lawton, Tristan Hume, Zac Hatfield-Dodds, Jared Kaplan, Jan Brauner, Samuel~R. Bowman, and Ethan Perez.
\newblock Measuring faithfulness in chain-of-thought reasoning, 2023.

\bibitem{mao2025earlystoppingchainofthoughtslarge}
Minjia Mao, Bowen Yin, Yu~Zhu, and Xiao Fang.
\newblock Early stopping chain-of-thoughts in large language models, 2025.

\bibitem{muennighoff-etal-2025-s1}
Niklas Muennighoff, Zitong Yang, Weijia Shi, Xiang~Lisa Li, Li~Fei-Fei, Hannaneh Hajishirzi, Luke Zettlemoyer, Percy Liang, Emmanuel Candes, and Tatsunori Hashimoto.
\newblock s1: Simple test-time scaling.
\newblock In Christos Christodoulopoulos, Tanmoy Chakraborty, Carolyn Rose, and Violet Peng, editors, {\em Proceedings of the 2025 Conference on Empirical Methods in Natural Language Processing}, pages 20275--20321, Suzhou, China, November 2025. Association for Computational Linguistics.

\bibitem{nathani2025mlgym}
Deepak Nathani, Lovish Madaan, Nicholas Roberts, Nikolay Bashlykov, Ajay Menon, Vincent Moens, Amar Budhiraja, Despoina Magka, Vladislav Vorotilov, Gaurav Chaurasia, et~al.
\newblock Mlgym: A new framework and benchmark for advancing ai research agents.
\newblock {\em arXiv preprint arXiv:2502.14499}, 2025.

\bibitem{openai2024learning}
{OpenAI}.
\newblock Learning to reason with llms.
\newblock \url{https://openai.com/index/learning-to-reason-with-llms/}, 2024.

\bibitem{paglieri2024balrog}
Davide Paglieri, Bart{\l}omiej Cupia{\l}, Samuel Coward, Ulyana Piterbarg, Maciej Wolczyk, Akbir Khan, Eduardo Pignatelli, {\L}ukasz Kuci{\'n}ski, Lerrel Pinto, Rob Fergus, et~al.
\newblock Balrog: Benchmarking agentic llm and vlm reasoning on games.
\newblock {\em arXiv preprint arXiv:2411.13543}, 2024.

\bibitem{quamar2025logitentropyadaptivestoppingheuristic}
Mohammad~Atif Quamar and Mohammad Areeb.
\newblock Logit-entropy adaptive stopping heuristic for efficient chain-of-thought reasoning, 2025.

\bibitem{ringel-etal-2025-learning}
Liran Ringel, Elad Tolochinsky, and Yaniv Romano.
\newblock Learning a continue-thinking token for enhanced test-time scaling.
\newblock In Kentaro Inui, Sakriani Sakti, Haofen Wang, Derek~F. Wong, Pushpak Bhattacharyya, Biplab Banerjee, Asif Ekbal, Tanmoy Chakraborty, and Dhirendra~Pratap Singh, editors, {\em Proceedings of the 14th International Joint Conference on Natural Language Processing and the 4th Conference of the Asia-Pacific Chapter of the Association for Computational Linguistics}, pages 3324--3345, Mumbai, India, December 2025. The Asian Federation of Natural Language Processing and The Association for Computational Linguistics.

\bibitem{schulman2017proximalpolicyoptimizationalgorithms}
John Schulman, Filip Wolski, Prafulla Dhariwal, Alec Radford, and Oleg Klimov.
\newblock Proximal policy optimization algorithms, 2017.

\bibitem{sel2025backtracking}
Bilgehan Sel, Dingcheng Li, Phillip Wallis, Vaishakh Keshava, Ming Jin, and Siddhartha~Reddy Jonnalagadda.
\newblock Backtracking for safety.
\newblock {\em arXiv preprint arXiv:2503.08919}, 2025.

\bibitem{sharma2025thinkjustenoughsequencelevel}
Aman Sharma and Paras Chopra.
\newblock Think just enough: Sequence-level entropy as a confidence signal for llm reasoning, 2025.

\bibitem{shridharalfworld}
Mohit Shridhar, Xingdi Yuan, Marc-Alexandre Cote, Yonatan Bisk, Adam Trischler, and Matthew Hausknecht.
\newblock Alfworld: Aligning text and embodied environments for interactive learning.
\newblock In {\em International Conference on Learning Representations}.

\bibitem{sun2025stopenoughadaptiveearlystopping}
Renliang Sun, Wei Cheng, Dawei Li, Haifeng Chen, and Wei Wang.
\newblock Stop when enough: Adaptive early-stopping for chain-of-thought reasoning, 2025.

\bibitem{tang2025concisehint}
Siao Tang, Xinyin Ma, Gongfan Fang, and Xinchao Wang.
\newblock Concisehint: Boosting efficient reasoning via continuous concise hints during generation.
\newblock {\em arXiv preprint arXiv:2506.18810}, 2025.

\bibitem{turpin2023languagemodelsdontsay}
Miles Turpin, Julian Michael, Ethan Perez, and Samuel~R. Bowman.
\newblock Language models don't always say what they think: Unfaithful explanations in chain-of-thought prompting, 2023.

\bibitem{wang2022scienceworld}
Ruoyao Wang, Peter Jansen, Marc-Alexandre C{\^o}t{\'e}, and Prithviraj Ammanabrolu.
\newblock Scienceworld: Is your agent smarter than a 5th grader?
\newblock In {\em Proceedings of the 2022 Conference on Empirical Methods in Natural Language Processing}, pages 11279--11298, 2022.

\bibitem{wen2025budgetthinker}
Hao Wen, Xinrui Wu, Yi~Sun, Feifei Zhang, Liye Chen, Jie Wang, Yunxin Liu, Yunhao Liu, Ya-Qin Zhang, and Yuanchun Li.
\newblock Budgetthinker: Empowering budget-aware llm reasoning with control tokens.
\newblock {\em arXiv preprint arXiv:2508.17196}, 2025.

\bibitem{wu2025effectively}
Tong Wu, Chong Xiang, Jiachen~T Wang, G~Edward Suh, and Prateek Mittal.
\newblock Effectively controlling reasoning models through thinking intervention.
\newblock {\em arXiv preprint arXiv:2503.24370}, 2025.

\bibitem{yang2025qwen3}
An~Yang, Anfeng Li, Baosong Yang, Beichen Zhang, Binyuan Hui, Bo~Zheng, Bowen Yu, Chang Gao, Chengen Huang, Chenxu Lv, et~al.
\newblock Qwen3 technical report.
\newblock {\em arXiv preprint arXiv:2505.09388}, 2025.

\bibitem{yang2025testtimepromptintervention}
Chenxu Yang, Qingyi Si, Mz~Dai, Dingyu Yao, Mingyu Zheng, Minghui Chen, Zheng Lin, and Weiping Wang.
\newblock Test-time prompt intervention, 2025.

\bibitem{yang2025dynamic}
Chenxu Yang, Qingyi Si, Yongjie Duan, Zheliang Zhu, Chenyu Zhu, Qiaowei Li, Minghui Chen, Zheng Lin, and Weiping Wang.
\newblock Dynamic early exit in reasoning models.
\newblock {\em arXiv preprint arXiv:2504.15895}, 2025.

\bibitem{yang2024swe}
John Yang, Carlos~E Jimenez, Alexander Wettig, Kilian Lieret, Shunyu Yao, Karthik Narasimhan, and Ofir Press.
\newblock Swe-agent: agent-computer interfaces enable automated software engineering.
\newblock In {\em Proceedings of the 38th International Conference on Neural Information Processing Systems}, pages 50528--50652, 2024.

\bibitem{yang2021safe}
Tsung-Yen Yang, Michael~Y Hu, Yinlam Chow, Peter~J Ramadge, and Karthik Narasimhan.
\newblock Safe reinforcement learning with natural language constraints.
\newblock {\em Advances in Neural Information Processing Systems}, 34:13794--13808, 2021.

\bibitem{yang2025step}
Xiao-Wen Yang, Xuan-Yi Zhu, Wen-Da Wei, Ding-Chu Zhang, Jie-Jing Shao, Zhi Zhou, Lan-Zhe Guo, and Yu-Feng Li.
\newblock Step back to leap forward: Self-backtracking for boosting reasoning of language models.
\newblock {\em arXiv preprint arXiv:2502.04404}, 2025.

\bibitem{yuan2025debuggymtextbasedenvironmentinteractive}
Xingdi Yuan, Morgane~M Moss, Charbel~El Feghali, Chinmay Singh, Darya Moldavskaya, Drew MacPhee, Lucas Caccia, Matheus Pereira, Minseon Kim, Alessandro Sordoni, and Marc-Alexandre Côté.
\newblock debug-gym: A text-based environment for interactive debugging, 2025.

\bibitem{zhang2025probinghidden}
Anqi Zhang, Yulin Chen, Jane Pan, Chen Zhao, Aurojit Panda, Jinyang Li, and He~He.
\newblock Reasoning models know when they're right: Probing hidden states for self-verification.
\newblock {\em arXiv preprint arXiv:2504.05419}, 2025.

\bibitem{zhang-etal-2025-steer}
Xingsheng Zhang, Luxi Xing, Chen Zhang, Yanbing Liu, Yifan Deng, Yunpeng Li, Yue Hu, and Chenxu Niu.
\newblock Can we steer reasoning direction by thinking intervention?
\newblock In Christos Christodoulopoulos, Tanmoy Chakraborty, Carolyn Rose, and Violet Peng, editors, {\em Findings of the Association for Computational Linguistics: EMNLP 2025}, pages 3888--3913, Suzhou, China, November 2025. Association for Computational Linguistics.

\bibitem{zhangbacktracking}
Yiming Zhang, Jianfeng Chi, Hailey Nguyen, Kartikeya Upasani, Daniel~M Bikel, Jason~E Weston, and Eric~Michael Smith.
\newblock Backtracking improves generation safety.
\newblock In {\em The Thirteenth International Conference on Learning Representations}.

\bibitem{zhao2025activation}
Zekai Zhao, Qi~Liu, Kun Zhou, Zihan Liu, Yifei Shao, Zhiting Hu, and Biwei Huang.
\newblock Activation control for efficiently eliciting long chain-of-thought ability of language models.
\newblock {\em arXiv preprint arXiv:2505.17697}, 2025.

\end{thebibliography}

\medskip


\appendix

\section{Embedded Answer Fixation}\label{app:ans_fix}
When initially attempting to train models to perform \bcreasoning, our initial approaches were pure prompting and RL fine-tuning.
However, we found both to fail due to the models' tendency to fixate on any previously mentioned answers within their reasoning trace.
We found models required SFT to properly adhere to the format of \bcreasoning where we observed an upside-down U curve in terms of performance, where early checkpoints perform strictly worse than the base model while failing to adhere to the \bcreasoning format; middle checkpoints both achieved proper formatting and comparable, if not greater, performance to the base model; and later checkpoints once again demonstrating a degradation in base model capabilities even while adhering to the \bcreasoning format.

We suspect the early collapse to be a symptom of the model initially fixating on the embedded answers during early training, resulting in a collapse of reasoning capabilities.
The middle checkpoints see a sharp increase in performance as the model learns to adhere to the \bcreasoning format without disrupting its natural reasoning.
The latter checkpoints collapse again as the model effectively begins to train on its own reasoning output.
The later checkpoint collapse is a studied effect that has evidence tracing back to GPT-2 \citep{herel2024collapseselftrainedlanguagemodels}.
For the early checkpoint result and question of why \ours cannot simply be elicited by SFT or RL, we perform a small scale experiment.

This experiment is performed post-hoc with a \bcreasoning model to prove that a base model will otherwise fixate to any answers included in its context rather than surface them and continue reasoning as done in \bcreasoning.
We take 100 context and reasoning trace pairs, generated by a \bcreasoning model.
For the reasoning, we truncate to the second \texttt{[answer]} block, feed the context and truncated reasoning back to a non-\bcreasoning model, and allow it to continue generating.
This advantages the base model towards generating a different answer than that exposed in the first \texttt{[answer]} block, as the stretch of reasoning after the first answer that leads the \bcreasoning to propse the second answer is explicitly provided to the base model.
Despite this, we found the base model would fixate and repeat the first answer \textbf{73\%} of the time.

During RL training, when the base model needs to instead construct a line of reasoning that leads it to a different answer itself, we saw this initial answer fixation almost always resulted in the second answer being a repeat of the first, resulting in non-convergence during training. 

A second motivation for avoiding RL training arises from the "monitorability tax" coined by Baker et al\citep{baker2025monitoringreasoningmodelsmisbehavior}.
RL training a model to perform \bcreasoning would explicitly require monitoring the reasoning trace to assign reward to ensure that the correct behavior is adhered to.
This both creates a dual optimization pressure, as the training objective would also require some pressure to ensure the model actually arrives at the correct final answer, as well as allowing for the potential of reward hacking.

\section{Cookingworld Data Filtering}\label{app:cooking_filter}
Cookingworld tasks involve the LLM acting as an agent to navigate a household setting, gather ingredients, and prepare a meal.
As ingredients are limited, failure can occur if an important ingredient is consumed or prepared in the wrong way, for example, eating a raw ingredient that should have been cooked.
If this occurs, the environment immediately terminates with a losing message.
The specific tasks and environment layouts used in \cite{cui2025talestextadventurelearning} are used as our test split with alternate recipes and environment layouts used as our train split.
A max step count of 20 is used as a tractable horizon that still gives the LLM sufficient attempts to make non-terminal mistakes in the environment.
A reasoning budget of 2048 tokens per step is allocated as higher token budgets resulted in minimal increases in performance for the base actor model.
Due the difficulty of the environment, both models failed to complete almost all of the included scenarios with a minimal prompt.
Following advice from \cite{cui2025talestextadventurelearning}, we include the contents of the \texttt{``help``} in the initial observation to aid the models in completing the given tasks.
We do not use the reasoning dead-end formulation in Cookingworld as due to the multi-step nature of the task, there is no ground-truth for if an action is 'incorrect' in the immediate context as it may lead to success or failure later on.
While task failure is possible in Cookingworld, for example eating a required ingredient for the meal when there are no backups, labeling all reasoning for these trajectories with the reasoning-dead-end formulation would result in a number of reasoning traces over identical or near-identical states being labeled with the reasoning dead-end labels as well as the successful reasoning trace labels, resulting in a significant amount of noise.
Furthermore, as the failure turns make up less than 5\% of the total data samples, we discard them to focus monitor evaluation in Cookingworld on efficiency in a multi-turn setting.

\section{Hazardworld changes}\label{app:hazard_changes}

We build on the HazardWorld benchmark of Yang et al. \citep{yang2021safe}, with modifications
grouped into four categories: task-tractability adjustments to make the
environment solvable by LLM agents, a text
observation and action interface, the scenario
suite used for train/test evaluation, and
benchmark-validity corrections to the original implementation.

\subsection{Task-tractability adjustments}
\label{app:tractability}

The original HazardWorld places its three pickup objects (Ball, Box, Key) and
hazard tiles independently. On large grids this frequently produces scenarios
in which an object is unreachable within a useful step budget, or in which the
constraint never interacts with the optimal trajectory. To make episodes both
tractable for LLM agents and informative about constraint adherence, we add
two placement mechanisms and expose hazard density as a hyperparameter.

\textbf{Chain object placement.}
When a ground-truth step budget \texttt{max\_steps\_gt} is supplied, objects are
placed sequentially along a chain Agent~$\to$~Ball~$\to$~Box~$\to$~Key, with
each leg constrained to lie between $2$ and
$\lfloor \texttt{max\_steps\_gt}/4 \rfloor$ Manhattan cells from the previous
waypoint. The lower bound of $2$ guarantees at least one intermediate cell on
each shortest path (used by hazard injection below); the upper bound bounds
the optimal trajectory length.

\textbf{Hazard-on-path injection.}
After object placement, we ensure that at least one tile of the avoid-hazard
type lies within the bounding box of every consecutive (waypoint, waypoint)
pair. If no such tile already exists, we place one on an empty cell, or
otherwise overwrite a non-avoid hazard. This guarantees that the agent must
reason about the constraint in order to traverse the chain. For
\texttt{HazardWorldSequential} with \texttt{isBefore=False}, we use the
dormant avoid object (the active \texttt{avoid\_obj} is \texttt{None} until
the trigger fires).

\textbf{Hazard density.}
We expose hazard density as a hyperparameter \texttt{sparsity}, with default
$0.75$ (raised from $0.50$ in the original). At $0.75$, $\sim\!61\%$ of
interior cells are hazardous and $\sim\!21\%$ are of the avoid type,
increasing the rate of constraint engagement per episode. We also reorder
placement in the Budgetary and Sequential subclasses to
agent~$\to$~objects~$\to$~hazards (matching Relational), so that chain
placement always operates on an empty interior.

\begin{table}[h]
\centering
\small
\begin{tabular}{lll}
\toprule
Parameter & Default & Effect \\
\midrule
\texttt{size}            & 13   & Grid side length (now passable to all subclasses) \\
\texttt{max\_steps\_gt}  & None & Ground-truth step budget; enables chain placement \\
\texttt{sparsity}        & 0.75 & Probability each interior cell is a hazard \\
\bottomrule
\end{tabular}
\caption{New environment parameters added to \texttt{HazardWorldBase}.}
\label{tab:hw-params}
\end{table}

\subsection{Text observation and action interface}
\label{app:text-interface}

We wrap the environment with a \texttt{HazardWorldTextWrapper} that produces
ASCII observations and accepts text actions, enabling LLM agents to play
without bespoke tokenisation.

\textbf{Grid encoding.}
Every cell is rendered as a fixed two-character code, chosen so that each
code is a single token in the Qwen3-8B tokenizer used in our experiments:
\texttt{\#\#}~Wall, \texttt{..}~Empty, \texttt{Lv}~Lava, \texttt{Gr}~Grass,
\texttt{Wa}~Water, \texttt{Bl}~Ball, \texttt{Bo}~Box, \texttt{Ky}~Key,
\texttt{??}~Unseen. The agent is rendered as
\texttt{>>}/\texttt{<<}/\texttt{vv}/\texttt{\^{}\^{}} in facing mode, where the agent can only move forward and must explicitly turn to change direction, or
\texttt{Ag} in cardinal mode. Single-token encoding avoids cross-cell token
boundaries that would otherwise force the model to spend reasoning capacity
on parsing the map.

\textbf{Observation and action options.}
\begin{itemize}
  \item \emph{Partial observability:} a configurable square window (default
        $3{\times}3$) around the agent; cells outside the window are rendered
        \texttt{??}. For window size $w$ the visibility radius is
        $r = (w-1)/2$.
  \item \emph{Fog of war} (optional): once-seen cells remain revealed for the
        rest of the episode.
  \item \emph{Cardinal movement} (optional): the four-action set
        \{\texttt{up}, \texttt{down}, \texttt{left}, \texttt{right}\}
        replaces the original turn-then-forward action set.
  \item \emph{Auto-pickup:} moving onto a pickupable object collects it
        automatically, matching the convention common in text-game baselines.
  \item \emph{Action parsing:} actions are extracted from
        \texttt{<action>$\,\cdot\,$</action>} tags in model output.
\end{itemize}

\subsection{Scenario suite}
\label{app:scenarios}

We release a fixed scenario file (\texttt{scenarios.json}) containing $30$
training and $30$ testing scenarios. Both splits share the same parameter
combinations but use disjoint random seeds, so test-time evaluation measures
generalisation to unseen layouts under matched constraints. Coverage is:
\begin{itemize}
  \item \textbf{Budgetary} (10): three avoid-hazard objects $\times$ hazard
        counts $\texttt{hc} \in \{1, 3, 5, 6\}$.
  \item \textbf{Sequential} (10): five with \texttt{isBefore=true} and five
        with \texttt{isBefore=false}, varying the
        (\texttt{first\_obj}, \texttt{avoid\_obj}) pair.
  \item \textbf{Relational} (10): three avoid-hazard objects $\times$
        \texttt{min\_dist} $\in \{1, 2\}$. The choice to omit
        \texttt{min\_dist} $\in \{0, 3\}$ is justified in
        \S\ref{app:corrections}.
\end{itemize}

\subsection{Benchmark-validity corrections}
\label{app:corrections}

While porting HazardWorld we identified three pre-existing issues that affect
the validity of comparisons drawn from the original implementation. We
document and correct them here.

\textbf{Non-deterministic scenario generation.}
The original \texttt{\_gen\_grid} mixes Gymnasium's seeded
\texttt{self.np\_random} (used for agent and object placement) with Python's
unseeded \texttt{random} module (used for constraint parameters and hazard
placement) and an unseeded \texttt{np.random.choice} call in
\texttt{make\_budgetary\_constraint}. Consequently, a fixed environment seed
does not reproduce the same scenario across runs. We apply standardization so that seed reproducability is ensured.

\textbf{Sequential constraint never deactivates.}
\texttt{HazardWorldSequential.step()} contains two bugs on the
\texttt{isBefore} branch in the original repository:
\begin{itemize}
  \item \texttt{curr\_cell == first\_obj} compares an object instance against
        an undefined local; the intended check is
        \texttt{curr\_cell.type == self.first\_obj}.
  \item \texttt{self.avoid\_obj == None} is a comparison rather than an
        assignment; the active constraint is therefore never deactivated
        after the trigger fires.
\end{itemize}
Both are corrected. Because the second bug silently inverts the semantics of
\texttt{isBefore=true} scenarios, prior numerical results obtained on the
original implementation in this regime should be interpreted with caution.

\textbf{Relational \texttt{min\_dist} is not enforced.}
The relational constraint nominally requires the agent to remain at least
\texttt{min\_dist} Manhattan-cells from the avoid object. However, the
original \texttt{step()} only registers a violation when the agent steps
\emph{onto} the avoid tile
(\texttt{curr\_cell.type == self.avoid\_obj}), never when it merely enters
the surrounding disk. Two consequences follow:
\begin{itemize}
  \item \texttt{min\_dist=0} is trivially always satisfied and therefore
        semantically empty.
  \item \texttt{min\_dist=2} and \texttt{min\_dist=1} produce identical
        runtime behavior despite being intended as different difficulty
        levels.
\end{itemize}
Additionally, under the partial-observation window of size $5$ (radius
$r = 2$) used in our experiments, an agent cannot see a hazard at Manhattan
distance $\geq 3$, so \texttt{min\_dist=3} is information-theoretically
infeasible. We therefore restrict our relational scenarios to
\texttt{min\_dist} $\in \{1, 2\}$. We further surface \texttt{min\_dist} in
the \texttt{info} dict and emit a warning when \texttt{min\_dist} exceeds the
partial-observation radius.

\textbf{Other.}
We additionally migrated the codebase from \texttt{gym} to \texttt{gymnasium}
(five-tuple \texttt{step}, \texttt{reset(seed, options)}, and the
\texttt{np\_random.integers} API) and enriched the \texttt{info} dict with
\texttt{constraint}, \texttt{max\_violations}, and a \texttt{full\_map}
rendering for downstream analysis. These do not affect the experimental
setup and are detailed in the released code.

\section{Oversight Module RL}\label{app:RL}
When optimizing the oversight module for efficiency, we treat the early stopping objective as a multi-turn task where the monitor is queried at each decision point and must decide whether reasoning should \texttt{[continue]} or \texttt{[stop]}.

As the outcome is based on the output of the monitor across all decision points at the time of termination, we use RL with a terminal reward to train the oversight module.
We compare a RL-training a monitor with the emergence of \texttt{[answer]} as decision points against a baseline where there is a decision point after each reasoning step.
In both cases, the \texttt{[answer]}, surfaced working answer, and \texttt{[continue]} blocks are parsed out due to fixation effect we saw when attempting to elicit a model to do \bcreasoning through prompting and RL.
For the plain thought baseline, we use the last formatted answer surfaced by \texttt{[answer]} to label decision points.
These experiments aim to identify if the decision points afforded by the surfacing of \texttt{[answer]} result in better RL training for a weaker monitor.
Due to the sheer length of the generated reasoning traces
In all experiments for this setting, we report percent of total possible token savings compared to an oracle and overall reward for validation. 
Metrics are reported in intervals of gradient update steps.

We use the verl-agent repository \citep{feng2025group} for RL fine-tuning.
We train using PPO \citep{schulman2017proximalpolicyoptimizationalgorithms} with a training batch size of 16, validation batch size of 32, and a learning rate of 1e-6.
There are two reward schema based on whether or not the final outcome of the reasoning trajectory is correct or not. 
In Cookingworld, all reasoning trajectories are labeled correct due to confounding factors in multi-turn environments discussed in Appendix \ref{app:cooking_filter}.
For correct trajectories, the monitor receives a +1 for stopping at the same answer the reasoning would ultimately terminate on with an scaling early stop bonus of up to .2 when terminating correctly.
Continuing past the natural end of reasoning results in a penalty of -1.
For incorrect trajectories (reasoning dead-ends), the monitor receives a +1 for stopping at any point during the trajectory with the same scaled early stop bonus.

For all monitor training scenarios, we observe the base model has a tendency to quickly collapse to always outputting 'continue' or 'stop' during early training.
To prevent this, we also include a small 'continue' reward to prevent early 'stop' collapse. 

\section{Oversight Module SFT}\label{app:SFT}
The safety objective represents a different paradigm for the oversight module and we use a different training method as a result.
For Hazardworld, at any decision point it doesn't matter whether previous actions in the reasoning trajectory were labeled as safe or unsafe.
This formulation is effectively a standard classification task.
The RL framing of the efficiency objective introduces noise in the safety objective as the safety of the current action surfaced by \texttt{[answer]} is independent of earlier action classifications.
The sole exception to this is when the earlier action and the current action are the \textit{same} action, however this still does not carry the same sequential correlation that necessitates the reward backpropagation in RL.
For this reason, we keep the same decision point formulation used to generate data samples but use SFT instead to train the monitor.
In all experiments for this setting, we report the accuracy and F1 score of safe and unsafe labeling. 
Metrics are reported in intervals of input tokens trained on.

We use the TRL framework for SFT. Fixed hyperparameters are used across both model families.

\section{Compute}\label{ref:compute}
We predict all experiments run end to end would take roughly a month on a cluster of A100 with 80GB of memory.

The primary bottleneck comes from the extended reasoning length allowed for the complex math problems in AIME. 
This necessitates a larger memory footprint for SFT and RL training.

Data generation for initial reasoning traces and subsequent \bcreasoning traces can be parallelized across all gpus for an 8B or 9B model relatively quickly.
Completions are far faster relatively, as the model is required to only generate a handful of tokens.

\section{Reporting}\label{app:reporting}
For AIME and Cookingworld baseline experiments, baselines results are reported as average over 3 seeds for both base model and for the \bcreasoning model. For Hazardworld, only one seed is used as the seeds are meant to represent an explicit scenario set verified by the authors.

For SFT and RL experiments for the monitor, only one seed is provided due to computational constraints in late stage testing. 
The shaded region in the RL experiments represents the standard deviation from the smoothing performed for graph presentation.

\section{Extended Societal Impacts}\label{app:extended}
This work is motivated by scalable oversight: making reasoning models easier to monitor and intervene on before final outputs are produced. In beneficial deployments, Behavior Cues could help reduce wasted inference compute, expose decision points for external oversight, and prevent unsafe commitments in constrained environments.

However, the same interface creates risks when reasoning is mediated through APIs or other systems that hide intermediate tokens from end users. Because Behavior Cues can be parsed or enforced without being surfaced in the final response, a platform or intermediary could use them to silently steer model reasoning without the user's awareness. This may be beneficial when used for disclosed safety interventions, but it also raises transparency and consent concerns if cue-based control is applied invisibly or for objectives misaligned with the user.

Finally, Behavior Cues should not be interpreted as deterministic guarantees of model behavior. LLM generation is inherently stochastic: our training procedure creates strong priors that anchor cue tokens to specific behaviors, but these associations are not guaranteed to hold on every sample or in every domain. This is reflected in the varying degrees of cue compliance observed across our experiments. We therefore view Behavior Cues as an interface that can improve monitorability and controllability, not as a replacement for careful validation, adversarial testing, and clear disclosure when reasoning-control mechanisms are deployed.


\end{document}